\documentclass{article}

\usepackage{arxiv}
\usepackage{float}
\usepackage[utf8]{inputenc}
\usepackage[T1]{fontenc}
\usepackage{hyperref}
\usepackage{url}
\usepackage{booktabs}
\usepackage{amsfonts}
\usepackage{amsmath}
\usepackage{nicefrac}
\usepackage{microtype}
\usepackage{graphicx}
\graphicspath{{./images/}}

\title{Gold Price Prediction Using Long Short-Term Memory and Multi-Layer Perceptron with Gray Wolf Optimizer}

\author{
  Hesam Taghipour \\
  College of Interdisciplinary Science and Technology\\
  University of Tehran\\
  Tehran, Iran\\
  \texttt{hesam.taghi.pour@ut.ac.ir} \\
  \And
  Alireza Rezaee \\
  College of Interdisciplinary Science and Technology\\
  University of Tehran\\
  Tehran, Iran\\
  \texttt{arrezaee@ut.ac.ir} \\
  \And
  Farshid Hajati \\
  School of Science and Technology\\
  University of New England\\
  Sydney, Australia\\
  \texttt{fhajati@une.edu.au} \\
}

\begin{document}
\maketitle

\begin{abstract}
The global gold market has long attracted financial institutions, banks, governments, funds, and micro-investors. Due to the inherent complexity and interdependence between important economic and political components, accurate forecasting of financial markets remains challenging. Therefore, providing a model that can accurately predict future market behavior is highly valuable.
In this paper, we present an artificial intelligence-based algorithm for daily and monthly gold price forecasting. Two long short-term memory (LSTM) networks are responsible for daily and monthly forecasting, and their outputs are integrated into a multilayer perceptron (MLP) network that provides the final forecast of the next day's prices. The algorithm forecasts the highest, lowest, and closing prices on both daily and monthly time frames. Based on these forecasts, we design a simple trading strategy for live market trading, under which the proposed model achieved a return of approximately 171\% in three months.
The number of internal neurons in each network is optimized by the Gray Wolf Optimizer (GWO) based on the minimum RMSE. The dataset spans 2010--2021 and includes macroeconomic indicators, energy markets, stock indices, and currency data of developed countries. Our proposed LSTM--MLP model predicts the daily closing price of gold with a mean absolute error (MAE) of \$0.21 and the next month's price with MAE of \$22.23.
\end{abstract}


\section{Introduction}

Gold has always been traded as a valuable commodity throughout history. With the expansion of banking systems, governments and financial institutions now play an important role in the global gold ounce market. Artificial intelligence algorithms have provided a platform for achieving accurate market predictions to help investors increase profits. Understanding complex relationships among economic variables enables better assessment of price changes, where disturbances in demand and supply cause significant price variations.

Kroner et al.\ proposed a model to predict commodity fluctuations \cite{Kroner1995}. Hochreiter and Schmidhuber introduced LSTM as a new recurrent neural network architecture that can learn long-term dependencies over more than 1000 discrete-time lags by propagating error through special memory cells \cite{Hochreiter1997, Gers2000}. Basheer and Hajmeer reviewed basic issues in the design and computation of artificial neural networks (ANNs) and proposed a generalized methodology for developing ANN-based models \cite{Basheer2000}.

Sjaastad studied the relationship between exchange rates and gold prices using theoretical and empirical methods, concluding that fluctuations in major exchange rates, especially the US dollar, are a primary source of instability in the world gold market \cite{Sjaastad2008}. Shafiee and Topal reviewed the world gold market, investigated the historical trend of gold prices, and compared it with parameters such as oil prices and global inflation, proposing a model to estimate gold prices for the next decade \cite{Shafiee2010}. Schmidhuber provided an overview of deep supervised, unsupervised, and reinforcement learning, as well as evolutionary computation and indirect search methods \cite{Schmidhuber2015}.

Sharma examined the effect of the consumer price index (CPI) from 54 countries on gold prices and found that gold price returns are predictable using CPI \cite{Sharma2016}. Mo et al.\ identified strong linkages among the gold market, US dollar, and crude oil market, and studied the effect of the global financial crisis on short-term and long-term memory properties of this volatility \cite{Mo2018}. Deng et al.\ applied deep reinforcement learning to financial signal representation and trading \cite{Deng2016}. Fischer and Krauss focused on LSTM applications to financial time series and compared LSTM with random forests and other deep networks within a trading strategy framework \cite{Fischer2018}. Mirjalili et al.\ developed the Gray Wolf Optimizer (GWO), a metaheuristic inspired by the hunting mechanism of grey wolves in nature \cite{Mirjalili2014}.

Sephton and Mann compared long-term fluctuations in gold and crude oil prices via nonlinear attractors \cite{Sephton2018}. Zakaria et al.\ used a whale optimization algorithm (WOA) to train an MLP neural network and compared it to PSO--NN, GA--NN, and GWO--NN baselines \cite{Alameer2019}. More broadly, deep and evolutionary models have been successfully deployed across diverse domains such as face and texture analysis \cite{Hajati2006FaceLocalization,Pakazad2006FaceDetection,Hajati2010PoseInvariant,Hajati2017DynamicTexture,Ayatollahi2015,Hajati2017Surface,AbdoliHajati2014,Shojaiee2014Palmprint,CremersACCV2014}, load forecasting \cite{Barzamini2012}, medical imaging and disease prediction \cite{Fiorini2019,Mahajan2024,Sopo2021DeFungi,Sadeghi2024COVID,Sadeghi2024ECG,Tavakolian2022FastCOVID,Tavakolian2023Readmission,Wang2022SoftwareImpacts}, and control and optimization \cite{Rezaee2010FIR,KarimiRezaee2017Helmholtz,MohamadzadeRezaee2017Antenna,Rezaee2017PID,Rezaee2017Penetrometer,Rezaee2017MPC,RezaeeGolpayegani2012,Rezaee2014FuzzyCloud,Gavagsaz2018LoadBalancing,Ramezani2024Drones}.

In this article, we estimate the global ounce price of gold in both the short and long term. Two LSTM networks are used for daily and monthly forecasting, and one MLP network combines the results from these two networks to produce a final daily forecast. The overall implementation steps are:
\begin{enumerate}
    \item Design the internal architecture of all networks. The algorithm consists of three main networks, each including three separate subnetworks for the high, low, and closing prices.
    \item Use the Gray Wolf Optimizer to determine the number of internal neurons in each layer of each network. The neuron counts for the highest, lowest, and closing prices in all three main networks are optimized separately.
    \item Define a simple trading strategy. After the networks make their predictions, the algorithm trades online in the gold market using the forecasted price ranges.
\end{enumerate}

\begin{figure}[H]
  \centering
  \includegraphics[width=0.9\linewidth]{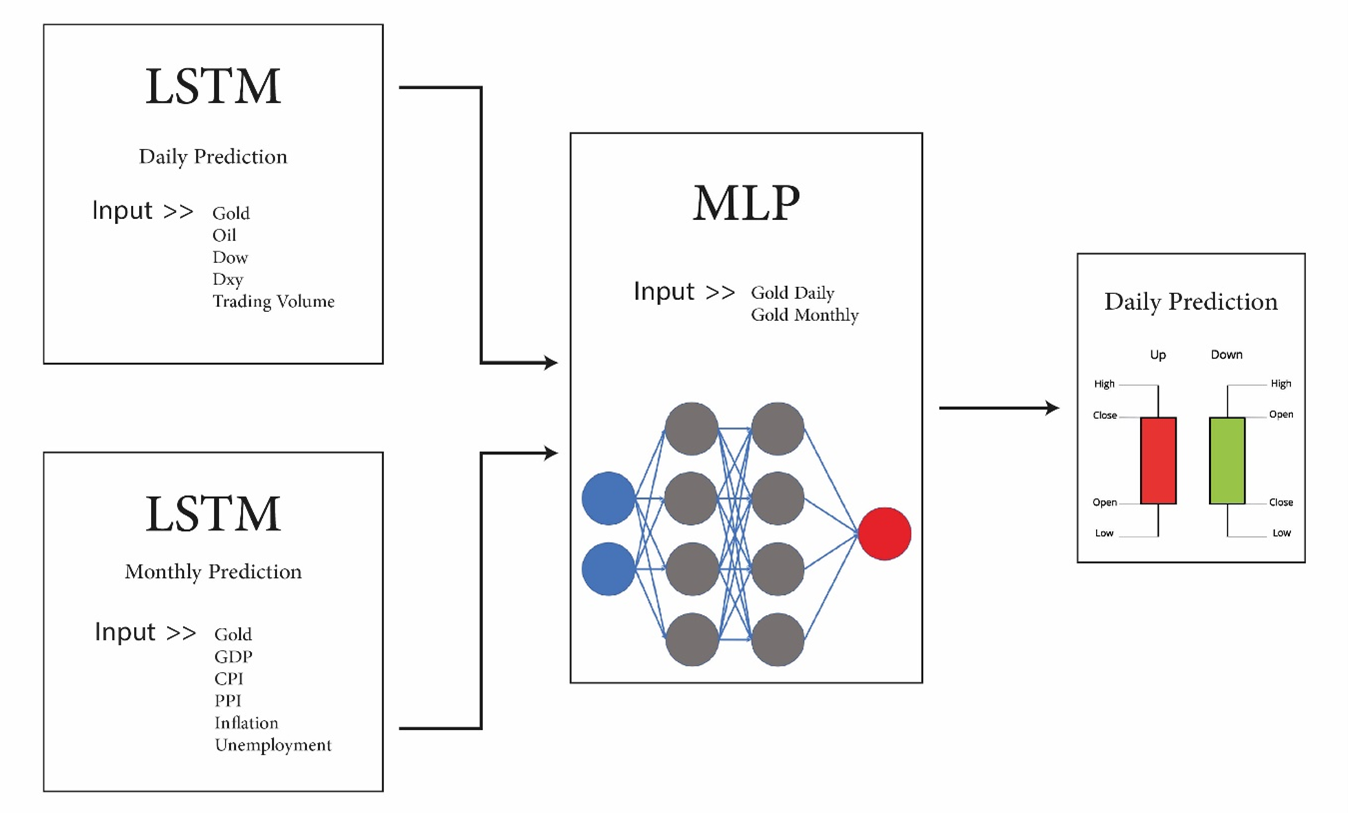}%
  \caption{Prediction steps of the proposed algorithm: data collection, preprocessing, LSTM-based daily and monthly forecasting, MLP fusion, and trading strategy generation.}
  \label{fig:algo_steps}
\end{figure}

The proposed algorithm therefore consists of nine separate networks, each trained on a different dataset. As a result, the suggested trades have reasonable targets based on these diverse predictions. The data includes daily and monthly gold prices, indicators from other markets, and macroeconomic data from the United States, as illustrated in Figure~\ref{fig:algo_steps}.

\section{Related Work}

Several studies have addressed gold price prediction using machine learning and metaheuristics. Li used a wavelet neural network (WNN) for gold price forecasting, optimized via an artificial bee colony (ABC) algorithm \cite{Li2014}. Aye et al.\ developed models based on six significant factors to forecast gold prices \cite{Aye2015}. Nassirtoussi et al.\ focused on intraday fluctuations of currency pairs in the FOREX market using text mining of financial headlines and breaking news \cite{Nassirtoussi2015}. Fan et al.\ developed an MLP network to predict coal prices with attention to nonlinear behavior and short-term features \cite{Fan2016}. Chong et al.\ predicted stock markets using principal component analysis, autoencoders, and restricted Boltzmann machines, building deep layers from seven feature sets \cite{Chong2017}.

Kanjilal and Ghosh studied the linkage between gold and crude oil prices using a vector error-correction model with a threshold effect \cite{Kanjilal2017}. Liu et al.\ forecast copper prices using a decision tree model with inputs including crude oil, natural gas, gold, silver, coffee prices, and the Dow Jones index \cite{Liu2017}. Hafezi and Akhavan proposed a BAT-optimized ANN model (BNN) for gold price prediction \cite{Hafezi2018}. Akbar et al.\ performed Bayesian analysis of dynamic linkages among gold price, stock prices, exchange rate, and interest rate \cite{Akbar2019}. Gheisari et al.\ leveraged genetic algorithms for job scheduling problems \cite{Gheisari2021}, demonstrating the effectiveness of evolutionary approaches in complex optimization. Zhang and Ci introduced a deep belief network (DBN) for gold price volatility forecasting \cite{Zhang2020}. Ingle and Deshmukh proposed an ensemble deep learning framework (EDLF-DP) that forecasts stock markets using online news data \cite{Ingle2021}. Jaquart et al.\ applied machine learning to short-term Bitcoin prediction using sub-hourly technical features \cite{Jaquart2021}.

In parallel, sophisticated pattern recognition and forecasting methods have been developed for other domains: offline signature and palmprint verification using geodesic or local-derivative patterns \cite{AbdoliHajati2014,Shojaiee2014Palmprint}, expression- and pose-invariant face recognition \cite{Ayatollahi2015,Hajati2010PoseInvariant,Hajati2017Surface,Hajati2017DynamicTexture}, short-term load forecasting \cite{Barzamini2012}, disease and therapy prediction \cite{Fiorini2019,Mahajan2024,Sadeghi2024ECG}, and COVID-19 diagnosis and software tools \cite{Sadeghi2024COVID,Tavakolian2022FastCOVID,Tavakolian2022SoftwareImpacts,Wang2022SoftwareImpacts}. These works demonstrate the robustness of deep and hybrid models in handling noisy, nonlinear, and high-dimensional data.

Our work contributes to this literature by:
\begin{itemize}
    \item Combining short-term (daily, technical) and long-term (monthly, macroeconomic) LSTM models with an MLP fusion network to produce next-day forecasts.
    \item Using GWO to optimize network architectures jointly across nine subnetworks (three price components $\times$ three main networks).
    \item Evaluating the model not only via error metrics but also via a concrete trading strategy in a live market environment.
\end{itemize}

\section{Methodology}

\subsection{Data}

In this study, data were collected from 1 January 2010 to the end of September 2021. To facilitate the training process, the data are standardized according to the approach in \cite{Shanker1996}. Let $x$ denote a raw value; a typical standardization is:
\[
z = \frac{x - \mu}{\sigma},
\]
where $\mu$ and $\sigma$ are the mean and standard deviation. After standardization, the data are divided into training, validation, and testing sets. Training and validation data are clustered using $k$-fold cross-validation.

Macroeconomic data (GDP, PPI, CPI, inflation, unemployment) are obtained from the US Federal Reserve website. Other data, including gold prices, Dow Jones index, oil prices, and gold trading volume, are collected from \url{https://finance.yahoo.com}. These datasets include the highest, lowest, and closing prices on daily and monthly bases. All steps are implemented in Python using standard data mining libraries.

The design---using parallel technical and fundamental datasets---is aligned with multimodal and multi-view integration in other domains, such as combining geometric and texture information for 3D deformable texture and face matching \cite{Hajati2017Surface,Ayatollahi2015,Hajati2010PoseInvariant} or integrating heterogeneous clinical and imaging biomarkers \cite{Fiorini2019,Sadeghi2024ECG}.

\subsection{LSTM Networks}

LSTM neural networks belong to the class of recurrent neural networks (RNN), designed to overcome the vanishing-gradient problem in standard RNNs \cite{Hochreiter1997,Gers2000}. In this study, two independent LSTM networks are designed.

The first LSTM network produces daily forecasts of the gold price. Its technical dataset includes daily gold price features (high, low, close), oil price, Dow Jones index, US dollar index, and daily trading volume of gold. These variables directly or indirectly influence gold prices.

The second LSTM network forecasts monthly gold prices using a fundamental dataset comprising macroeconomic parameters such as GDP, CPI, PPI, unemployment rate, and inflation rate.

In both LSTM networks, a leaky ReLU activation function is used, typically
\[
f(x) = 
\begin{cases}
x, & x \ge 0,\\
\alpha x, & x < 0,
\end{cases}
\]
with a small leakage coefficient $\alpha$ (e.g., $0.01$). This prevents ``dead'' neurons by allowing a small gradient when the unit is not active. A dropout layer with dropout rate of 0.01 is used to reduce overfitting. Early stopping is employed with a patience of 5 epochs, restoring the model weights associated with the minimum validation loss.

Both LSTM models are compiled using the Adam optimizer, mean squared error (MSE) as the loss function, and mean absolute error (MAE) as an evaluation metric, consistent with prior financial LSTM studies \cite{Fischer2018}. Similar loss and regularization strategies have been effective in other temporal and spatio-temporal deep models for face and medical data \cite{Hajati2017DynamicTexture,Sadeghi2024ECG}.

\subsection{MLP Fusion Network}

Multilayer perceptron (MLP) networks consist of input and output layers with one or more hidden layers. With sufficient hidden units, MLPs are universal approximators for continuous functions \cite{HechtNielsen1990,Hornik1989}. In this work, the MLP fusion network combines the outputs of the daily and monthly LSTM models to provide the final next-day prediction.

The input to the MLP comprises:
\begin{itemize}
    \item Short-term (daily) forecasts from the first LSTM,
    \item Long-term (monthly) forecasts from the second LSTM.
\end{itemize}
Dropout is set to 0.01, and leaky ReLU is again used as the activation function with a leakage coefficient of 0.01. As with the LSTM models, four folds are used for training and validation, and the test set is held out for final evaluation. The MLP is compiled using Adam, with MSE as the loss and MAE as a metric, and early stopping patience is set to 5.

\subsection{Gray Wolf Optimizer}

To determine the most appropriate number of neurons in each layer of the networks, we use the Gray Wolf Optimizer (GWO) \cite{Mirjalili2014}. Each candidate solution (wolf) encodes the number of neurons in the three layers of a given subnetwork. GWO simulates the hunting behavior of grey wolves, with the best solutions denoted as $\alpha$, $\beta$, and $\delta$ wolves guiding the others.

All three main networks (two LSTMs and one MLP) have three layers for each of the three price components (high, low, close), leading to nine subnetworks in total. GWO is used to optimize the neuron counts in each layer for each of these nine subnetworks. The fitness function is the RMSE over a validation set; GWO searches for the neuron configuration that minimizes RMSE.

The GWO algorithm in this study uses:
\begin{itemize}
    \item Number of wolves in the herd: 5.
    \item Number of iterations: 10,
    \item Maximum neurons per layer: 1024,
    \item Minimum neurons per layer: 2,
    \item Number of features per wolf: 3 (one per layer).
\end{itemize}

\begin{figure}[H]
  \centering
  \includegraphics[width=0.95\linewidth]{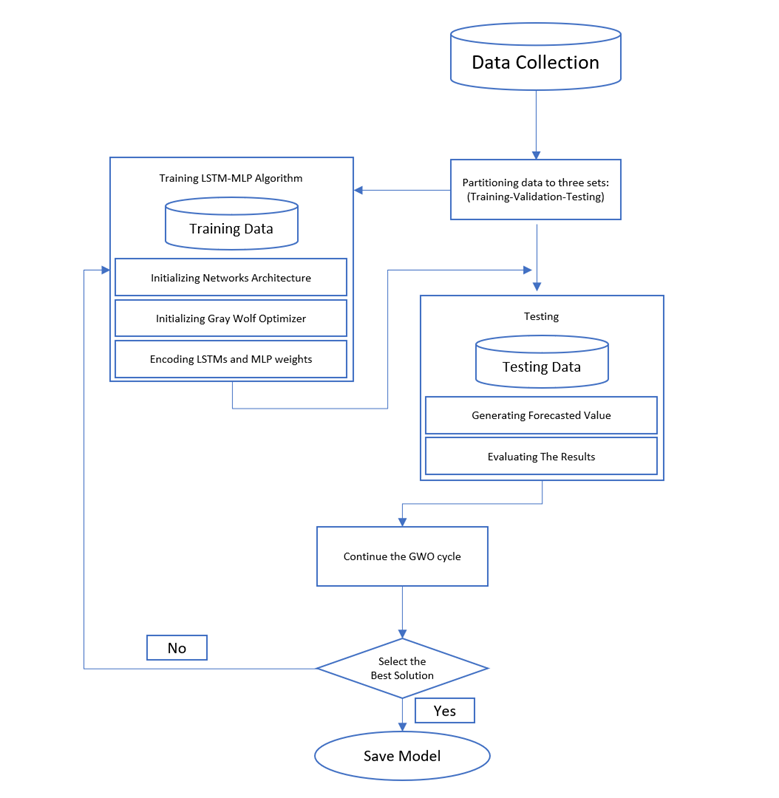}%
  \caption{Architecture of the proposed LSTM--MLP algorithm. Two LSTM networks handle daily and monthly forecasts, respectively. Their outputs feed into an MLP fusion network that produces final predictions for high, low, and close prices.}
  \label{fig:lstm_mlp_arch}
\end{figure}

Figure~\ref{fig:lstm_mlp_arch} shows the overall LSTM--MLP architecture.

\subsection{Benchmark Models}

We compare the proposed model against several benchmark models commonly used in financial time series forecasting:

\begin{itemize}
    \item \textbf{Feedforward Neural Network (FNN):} A standard multilayer feedforward network used for regression \cite{Hornik1989}.
    \item \textbf{Backpropagation Neural Network (BPNN):} A classical ANN trained via gradient backpropagation, computing gradients of the loss function and updating weights iteratively \cite{Basheer2000}.
    \item \textbf{Radial Basis Function Network (RBF):} Networks using radial basis functions (e.g., Gaussian) as activation functions \cite{Basheer2000}.
    \item \textbf{General Regression Neural Network (GRNN):} A regression model based on kernel estimators that provides smooth interpolation \cite{Specht1991}.
    \item \textbf{Adaptive Network-Based Fuzzy Inference System (ANFIS):} A neuro-fuzzy system that maps inputs to outputs using fuzzy rules and hybrid learning \cite{Jang1993}.
    \item \textbf{Empirical Mode Decomposition (EMD):} A decomposition technique using intrinsic mode functions and Hilbert transforms to analyze nonlinear, non-stationary time series \cite{Huang1998}, used as a preprocessing step for networks such as FNN and ALNN in EMD--FNN--ALNN.
    \item \textbf{Bat-Inspired Neural Network (BNN):} An ANN whose weights are optimized using the BAT algorithm inspired by bat echolocation \cite{Yang2010,Hafezi2018}.
    \item \textbf{Improved Empirical Mode Decomposition (IEMD) Networks:} Enhanced EMD-based architectures, such as IEMD--BPNN--PMR, using intrinsic mode functions combined with BPNN for time series classification and regression \cite{Zhou2012}.
\end{itemize}

These models provide a comprehensive set of baselines spanning conventional neural, neuro-fuzzy, and metaheuristic-optimized architectures.

\subsection{Trading Strategy}

The forecasts produced by the algorithm include the daily high, low, and closing prices. These are used to construct candlestick patterns for the next day \cite{Marshall2006}. Based on these patterns, a simple trading strategy is defined \cite{Conrad1998}:

\begin{itemize}
    \item The predicted closing price determines the direction (``long'' or ``short'').
    \item Predicted highs and lows determine suitable entry levels and profit targets.
    \item Orders are placed using pending order types (Buy Stop, Buy Limit, Sell Stop, Sell Limit) in MetaTrader.
    \item The Take-Profit (TP) is set at the predicted target level, and the Stop-Loss (SL) is set to one-third of TP distance to enforce a 3:1 reward-to-risk ratio.
    \item For each trade, 5\% of the free margin is risked. With a leverage of 1:100, the lot size is determined by:
    \[
    \text{Lot} = f(\text{FreeMargin}, \text{Leverage}, \text{SL}),
    \]
    where the exact function depends on broker specifications.
\end{itemize}

All trades are executed on a demo account with an initial balance of \$1{,}000. The overall trading performance over a three-month period is then measured.

\section{Results}

\subsection{Architecture Optimization}

The GWO algorithm is used to determine the optimal number of neurons in each of the three layers for each of the nine subnetworks. Table~\ref{tab:neurons} reports the final neuron configurations, where each entry denotes ``layer1--layer2--layer3''.

\begin{table}[H]
  \centering
  \caption{Optimal number of neurons in each network (first, second, and third layer).}
  \label{tab:neurons}
  \begin{tabular}{lccc}
    \toprule
    Network & High & Close & Low \\
    \midrule
    Daily LSTM   & 673--851--810 & 90--36--69   & 53--343--97 \\
    Monthly LSTM & 713--289--533 & 441--581--671 & 386--461--381 \\
    Final MLP    & 479--440--393 & 317--327--362 & 105--275--319 \\
    \bottomrule
  \end{tabular}
\end{table}

\begin{figure}[H]
  \centering
  \includegraphics[width=\linewidth]{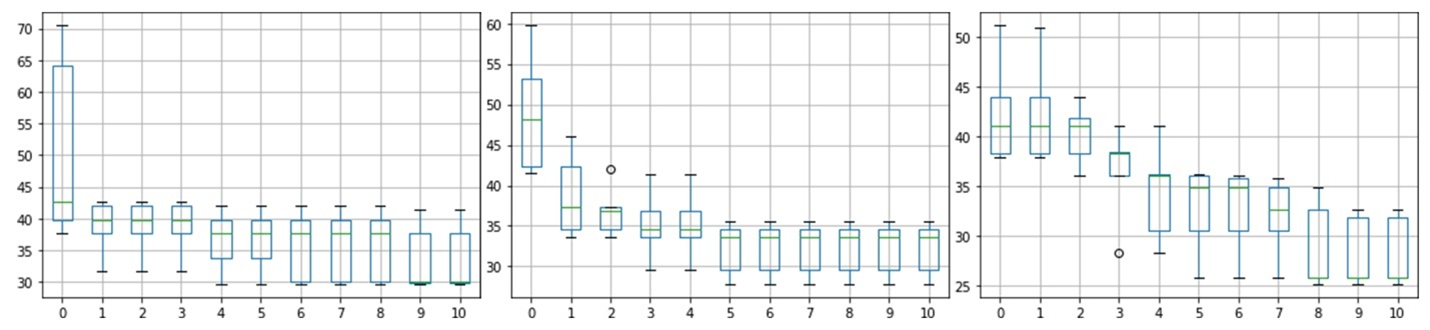}%
  \caption{Optimization of neuron counts for the daily LSTM subnetworks: High (left), Close (middle), Low (right).}
  \label{fig:gwo_daily}
\end{figure}

\begin{figure}[H]
  \centering
  \includegraphics[width=\linewidth]{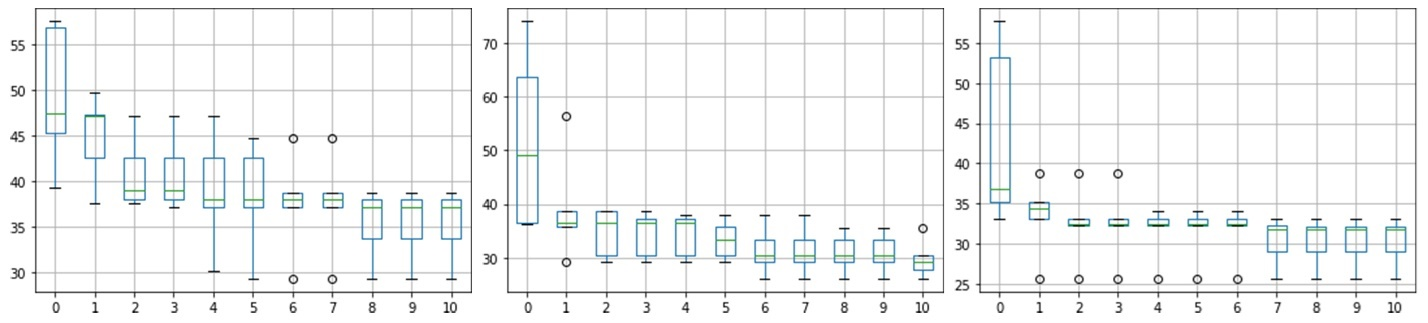}%
  \caption{Optimization of neuron counts for the monthly LSTM subnetworks: High (left), Close (middle), Low (right).}
  \label{fig:gwo_monthly}
\end{figure}

\begin{figure}[H]
  \centering
  \includegraphics[width=\linewidth]{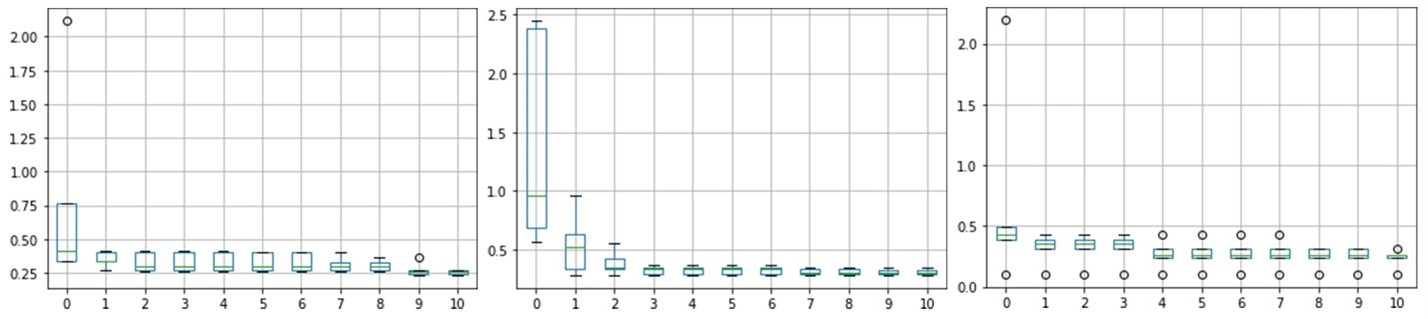}%
  \caption{Optimization of neuron counts for the MLP fusion subnetworks: High (left), Close (middle), Low (right).}
  \label{fig:gwo_mlp}
\end{figure}

Figures~\ref{fig:gwo_daily}, \ref{fig:gwo_monthly}, and \ref{fig:gwo_mlp} illustrate the optimization process for the first, second, and third networks, respectively, showing RMSE versus iteration for high, close, and low subnetworks.

\subsection{Prediction Accuracy}

After fixing the architectures, the networks are trained and evaluated on the test set (approximately 10\% of the total data). Prediction errors are reported using RMSE and MAE. Table~\ref{tab:errors} summarizes the results for daily and monthly forecasts, with units in US dollars.

\begin{table}[H]
  \centering
  \caption{Prediction errors for Low, Close, and High prices on daily and monthly time frames.}
  \label{tab:errors}
  \begin{tabular}{lcccccc}
    \toprule
    & \multicolumn{3}{c}{Daily} & \multicolumn{3}{c}{Monthly} \\
    \cmidrule(lr){2-4} \cmidrule(lr){5-7}
    Metric & Low & Close & High & Low & Close & High \\
    \midrule
    RMSE & 0.24 & 0.28 & 0.23 & 25.65 & 26.07 & 29.31 \\
    MAE  & 0.19 & 0.21 & 0.15 & 21.74 & 22.23 & 22.98 \\
    \bottomrule
  \end{tabular}
\end{table}

These results indicate that the proposed LSTM--MLP model achieves high accuracy in predicting the daily and monthly gold prices, with particularly low errors for the daily close and monthly close series. As argued by Willmott and Matsuura \cite{Willmott2005}, MAE is often a more interpretable and robust measure than RMSE for average model performance. Similar error scales and improvements over baselines have been reported in other forecasting applications, including load forecasting and disease prediction \cite{Barzamini2012,Mahajan2024,Fiorini2019}.

\subsection{Trading Performance}

\begin{figure}[H]
  \centering
  \includegraphics[width=0.9\linewidth]{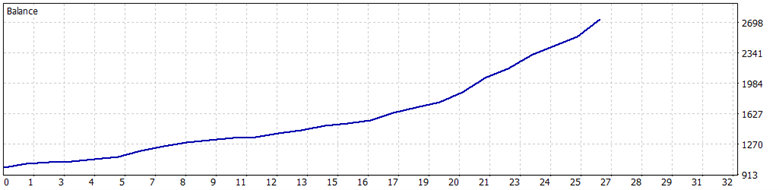}%
  \caption{Result of trading using the model's daily signals over three months. The strategy achieved an approximate return of 172--173\%.}
  \label{fig:trading}
\end{figure}

Figure~\ref{fig:trading} shows the equity curve for algorithmic trading based on the model's signals using the simple daily strategy described earlier, over a three-month period on a demo account starting from \$1{,}000.

The strategy produced a return of approximately 172--173\% over three months, demonstrating that the model's forecasts, when embedded in a consistent trading framework, can lead to profitable decisions. This stands in contrast to much of the prior literature, which reports only prediction errors without assessing live trading performance \cite{Li2014,Aye2015,Hafezi2018,Zhang2020}. The focus on end-to-end pipelines---from prediction to action---parallels recent work in medical decision support and resource allocation \cite{Fiorini2019,Tavakolian2023Readmission,Sadeghi2024COVID,Tavakolian2022022}.

\subsection{Comparison with Benchmark Models}

Tables~\ref{tab:daily_bench} and \ref{tab:monthly_bench} compare our daily and monthly forecasts with several benchmark models.

\begin{table}[H]
  \centering
  \caption{Comparison of daily forecasts (MAE) with benchmark models.}
  \label{tab:daily_bench}
  \begin{tabular}{lcc}
    \toprule
    Model & MAE (Daily) & Reference \\
    \midrule
    BPNN              & 9.82 & \cite{Basheer2000} \\
    FNN               & 8.97 & \cite{Hornik1989} \\
    EMD--FNN--ALNN    & 3.35 & \cite{Huang1998,Zhou2012} \\
    IEMD--BPNN--PMR   & 2.17 & \cite{Zhou2012} \\
    Proposed LSTM--MLP & \textbf{0.21} & This work \\
    \bottomrule
  \end{tabular}
\end{table}

\begin{table}[H]
  \centering
  \caption{Comparison of monthly forecasts (RMSE) with benchmark models.}
  \label{tab:monthly_bench}
  \begin{tabular}{lcc}
    \toprule
    Model & RMSE (Monthly) & Reference \\
    \midrule
    RBF    & 124.36 & \cite{Basheer2000} \\
    MLP    & 74.19  & \cite{Hornik1989} \\
    ANFIS  & 29.48  & \cite{Jang1993} \\
    BNN    & 21.26  & \cite{Hafezi2018,Yang2010} \\
    Proposed LSTM--MLP & 26.07  & This work \\
    \bottomrule
  \end{tabular}
\end{table}

The proposed model significantly outperforms classical neural and EMD-based models in daily forecasting and achieves competitive performance for monthly forecasts relative to advanced metaheuristic-optimized networks.

\section{Discussion}

We proposed a technique that combines LSTM and MLP neural networks, together with GWO-based architecture optimization, to predict the global ounce price of gold on daily and monthly time frames. The algorithm integrates macroeconomic data and information from related markets, providing a richer feature space than many previous studies \cite{Shafiee2010,Sharma2016,Mo2018,Sephton2018,Alameer2019,Zhang2020}.

Our experimental results show that:
\begin{itemize}
    \item The daily forecast achieves MAE and RMSE of approximately \$0.21 and \$0.28 for the closing price, and as low as \$0.15 and \$0.23 for the low price.
    \item Monthly forecasts achieve MAE and RMSE of approximately \$22.23 and \$26.07 for the closing price.
    \item In live demo trading, a simple strategy using the model's forecasts achieved a return of about 173\% over three months.
\end{itemize}

Notably, many earlier works evaluated their models only via error metrics, without reporting performance in a trading setting. Our analysis bridges this gap by showing that the forecasts can translate into profitable algorithmic trading, building on broader insights from algorithmic trading research \cite{Nuti2011,Oliveira2017} and high-performance implementations in other domains \cite{Rezaee2014FuzzyCloud,Gavagsaz2018LoadBalancing,BarolliAINA2024,BarolliBWCCA2019,BarolliWAINA2019}.

Economic cycles and regime shifts are important topics that could further enhance this study. Grinin et al.\ discussed how economic cycles and crises can emerge from random discrete events \cite{Grinin2016}, which suggests that incorporating regime-switching or cycle-aware components into the model might improve robustness. Moreover, news and social-media signals have been shown to influence market fluctuations \cite{Oliveira2017,Nassirtoussi2015,Ingle2021}, and could be integrated as additional input streams in future versions of the model.

In addition, methodological ideas from other successful applications---such as derivative-based dynamic texture descriptors \cite{Hajati2017DynamicTexture}, geodesic-based representations \cite{Hajati2017Surface,AbdoliHajati2014}, and attention-based clinical models \cite{Fiorini2019, Mahajan2024,Sadeghi2024ECG}---could inspire more interpretable or robust architectures for financial forecasting.

\section{Conclusion}

In this paper, we predicted the global gold ounce price using LSTM and MLP neural networks, with the Gray Wolf Optimizer used to enhance model performance by optimizing neuron counts. Forecasts were performed on daily and monthly time frames and included high, low, and closing prices. The model leveraged macroeconomic indicators, energy prices, stock indices, and other market variables.
Gray wolves optimized the internal architectures of nine subnetworks, which were then trained to predict gold prices. The algorithm achieved MAE and RMSE of approximately \$0.15 and \$0.23 (daily low) and \$22.98 and \$29.31 (monthly high). These errors are favorable compared with benchmark models, and the trading performance was strong, with an approximate 173\% return over three months on a demo account.
For future work, additional data sources such as sentiment indices, microblog data, and web search trends \cite{Oliveira2017,Jaquart2021} could be incorporated. Furthermore, extending the model to other precious metals or commodities and integrating more sophisticated risk-management and portfolio-allocation strategies would be natural next steps. Given the success of related AI frameworks in other domains---from smart grids and control \cite{Barzamini2012,Rezaee2017MPC,RezaeeGolpayegani2012,Rezaee2010FIR,KarimiRezaee2017Helmholtz,MohamadzadeRezaee2017Antenna,Taghvaee2014Metamaterial} to medical imaging and decision support \cite{Fiorini2019,Sadeghi2024COVID,Sadeghi2024ECG,Sopo2021DeFungi,Tavakolian2022FastCOVID,Tavakolian2023Readmission,Shahramian2013Leptin,Shahramian2013Troponin}, we anticipate that similar hybrid and interpretable designs will further advance financial time-series forecasting.

\bibliographystyle{unsrt}

\end{document}